\documentclass[10pt,twocolumn,letterpaper]{article}

\usepackage[pagenumbers]{cvpr} 

\usepackage{graphicx}
\usepackage{capt-of}
\usepackage{amsmath,amssymb,amsfonts,dsfont,bm,bbm,mathrsfs,pifont}
\usepackage{booktabs,multirow,adjustbox}
\usepackage[pagebackref,breaklinks,colorlinks]{hyperref}
\usepackage[capitalize]{cleveref}  
\usepackage{arydshln}
\usepackage{algorithm}
\usepackage{verbatim}
\usepackage{listings}
\usepackage{algpseudocode}

\usepackage{enumitem}
\usepackage[accsupp]{axessibility} 

\crefname{section}{Sec.}{Secs.}
\Crefname{section}{Section}{Sections}
\crefname{table}{Tab.}{Tabs.}
\Crefname{table}{Table}{Tables}
\crefname{figure}{Fig.}{Figs.}
\Crefname{figure}{Figure}{Figures}
\crefname{equation}{Eq.}{Eqs.}
\Crefname{equation}{Equation}{Equations}
\def\cvprPaperID{****}
\def\confName{CVPR}
\def\confYear{2022}

\begin{document}

\title{Improving GAN Equilibrium by Raising Spatial Awareness}

\author{Jianyuan Wang$^{1,2}$ \quad Ceyuan Yang$^1$  \quad Yinghao Xu$^1$ \quad Yujun Shen$^1$ \quad Hongdong Li$^2$ \quad Bolei Zhou$^3$ \\
$^1$The Chinese University of Hong Kong \ \ $^2$The Australian National University \\
$^3$University of California, Los Angeles
}

\maketitle

\begin{abstract}

The success of Generative Adversarial Networks (GANs) is largely built upon the adversarial training between a generator ($G$) and a discriminator ($D$).
They are expected to reach a certain equilibrium where $D$ cannot distinguish the generated images from the real ones.
However, such an equilibrium is rarely achieved in practical GAN training, instead, $D$ almost always surpasses $G$.
We attribute one of its sources to the information asymmetry between $D$ and $G$.
We observe that $D$ learns its own visual attention when determining whether an image is real or fake, but $G$ has no explicit clue on which regions to focus on for a particular synthesis.
To alleviate the issue of $D$ dominating the competition in GANs, we aim to raise the spatial awareness of $G$. 
Randomly sampled multi-level heatmaps are encoded into the intermediate layers of $G$ as an inductive bias.
Thus $G$ can purposefully improve the synthesis of certain image regions.
We further propose to align the spatial awareness of $G$ with the attention map induced from $D$.
Through this way we effectively lessen the information gap between $D$ and $G$. 
Extensive results show that our method pushes the two-player game in GANs closer to the equilibrium, leading to a better synthesis performance.
As a byproduct, the introduced spatial awareness facilitates interactive editing over the output synthesis. Demo video and code are available at \url{https://genforce.github.io/eqgan-sa/}.
\end{abstract}

\section{Introduction}\label{sec:intro}

\begin{figure}[t]
    \centering
    \includegraphics[width=0.95\linewidth]{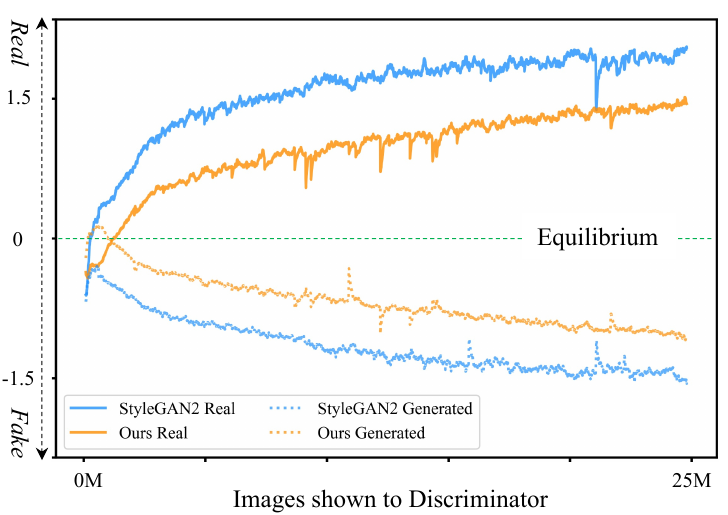}
    \vspace{-5pt}
    \caption{
    \textbf{Discriminator's output scores for real and generated samples during training.} A higher number indicates more realistic. We use StyleGAN2~\cite{stylegan2} as the baseline and implement the proposed method over it on the LSUN Cat dataset. For a clear comparison, we report the minimum of the scores for real samples and the maximum for generated samples, which are supposed to be close to each other. However, the curve for the real is always above the curve for the generated. Our method can reduce the gap toward the equilibrium as well as improve the synthesis quality.}
    \label{Fig:Logits_Cat}
    \vspace{-10pt}
\end{figure}

Generative Adversarial Network (GAN) has made huge progress toward image synthesis~\cite{goodfellow2014generative, dcgan, bigGAN, stylegan, stylegan2}.
GAN is formulated as a two-player game between a generator ($G$) and a discriminator ($D$)~\cite{goodfellow2014generative}, where $G$ targets at reproducing the distribution of observed data through synthesizing new samples, and $D$ competes with $G$ by distinguishing the generated images from the real ones.
In principle, they are expected to reach an equilibrium where $D$ cannot tell the real and fake images apart~\cite{goodfellow2014generative, wgan}.

In practice, it turns out to be difficult to achieve such an equilibrium when training modern GAN variants~\cite{bigGAN, stylegan, stylegan2, stylegan2-ada}, despite their appealing synthesis quality. 
Taking StyleGAN2~\cite{stylegan2} as an example, $D$ almost always assigns a higher score to real images than to fake ones throughout the \textit{entire} training process, as the blue lines shown in \cref{Fig:Logits_Cat}.
It suggests that $D$ can easily beat $G$ in the competition.
Increasing the model capacity of $G$ barely helps mitigate this issue~\cite{bigGAN}.
From this point of view, we can hold that $G$ fails to fool $D$ even after the model converges, leaving a gap between the real and synthesized distributions.
Such a disequilibrium remains much less explored in recent years, in spite of the rapid development of new GAN models.

\begin{figure}[t]
\begin{center}
    \includegraphics[width=0.92\linewidth]{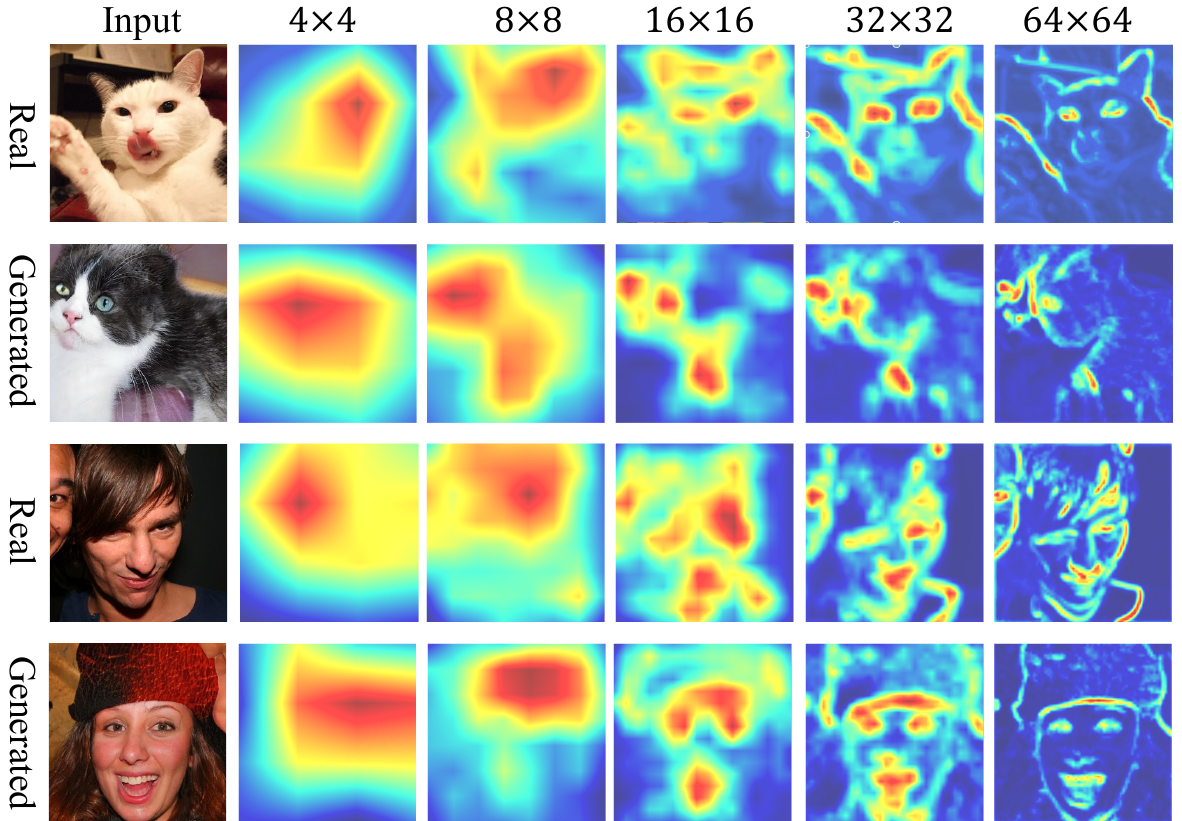}
    \caption{\textbf{Spatial visual attention at the intermediate layers of the discriminator,} visualized by GradCAM. A bright color indicates a strong contribution to the final score. `$64\times64$' indicates being upsampled from a $64\times64$ feature map. The samples are the real images and the images generated by StyleGAN2~\cite{stylegan2}.}    
    \label{Fig:Hie}
\end{center}
\vspace{-15pt}
\end{figure}

To investigate the source of the aforementioned disequilibrium, we analyze the behaviour of $D$ to figure out its advantage over $G$.
With the help of GradCAM~\cite{gradcam} as a neural network interpretation tool, we visualize the intermediate feature maps produced by $D$.
As shown in \cref{Fig:Hie}, given a target image, either real or generated, $D$ holds its own visual attention, even it is \textit{merely} learned from a bi-classification task (\textit{i.e.}, differentiating the real and fake domains).
In other words, $D$ is aware of which flawed regions to pay attention to when making the real/fake decision.
Such an attentive property eases its competition to $G$ because $D$ can simply focus on the regions that are poorly synthesized by $G$.
On the contrary, to produce an image, $G$ takes a randomly sampled latent code as the input and has no explicit clue about which regions to focus on, let alone knowing the spatial preference of $D$.
Due to such an information asymmetry, $G$ may omit some important cues picked by $D$ and hence get defeated in the two-player game. 

In this paper, we propose a training method, termed as \textbf{EqGAN-SA}, to improve the \textbf{Eq}uilibrium of \textbf{GAN} through raising \textbf{S}patial \textbf{A}wareness.
Concretely, we strive to lessen the information asymmetry between $G$ and $D$ by raising the \textit{spatial awareness} of $G$.
We design a hierarchical heatmap sampling strategy to match the coarse-to-fine synthesis mechanism~\cite{stylegan, higan}.
The sampled multi-level heatmaps are integrated into the per-layer feature maps of $G$.
Meanwhile, to make sure $G$ utilize the input heatmap adequately, we involve $D$ as a regularizer to spatially supervise the generation process, through aligning the spatial awareness of $G$ with the visual attention induced from $D$.

We evaluate the proposed method on various datasets.
As the orange lines shown in \cref{Fig:Logits_Cat}, we push the competition between $G$ and $D$ closer to the equilibrium compared to the baseline method~\cite{stylegan2}.
Consequently, our EqGAN-SA learns a distribution that is more identical to the real one, leading to a substantial improvement in the synthesis performance.
For example, with Fréchet Inception Distance (FID)~\cite{heusel2017gans} as the metric, we improve the baseline from $3.66$ to $2.96$ on FFHQ dataset~\cite{stylegan} under $256\times256$ resolution.
In addition, we can achieve interactive editing over the output image by altering the spatial heatmaps fed into $G$. 

\section{Analyzing GAN Equilibrium}\label{sec:analysis}

\begin{figure*}[t]
    \centering
    \includegraphics[width=0.95\linewidth]{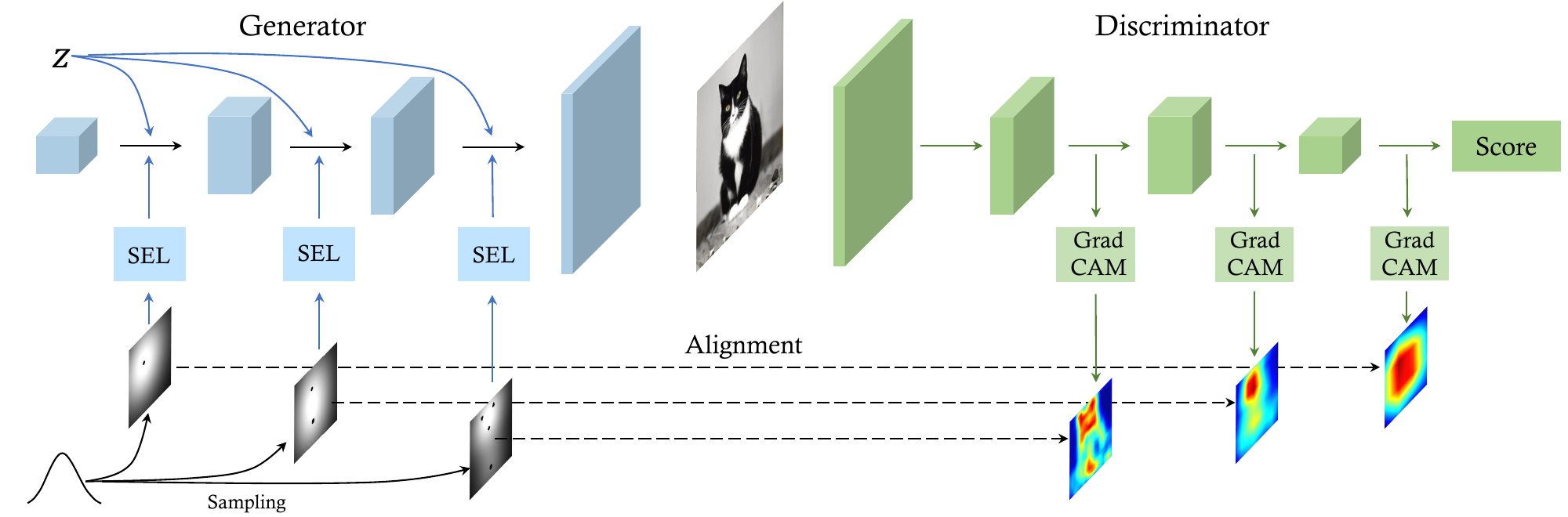}
    \vspace{-5pt}
    \caption{\textbf{Illustration of EqGAN-SA.} We conduct spatial encoding in $G$ and align its spatial awareness with $D$ attention maps. Specifically, we randomly sample spatial heatmaps and encode them into $G$ via the spatial encoding layer (SEL). To implement the alignment during training, we calculate $D$ attention maps over the generated samples via GradCAM. 
    }
    \label{Fig:Landscape}
\vspace{-10pt}
\end{figure*}

Although GANs~\cite{goodfellow2014generative,wgan,BigBiGAN,stylegan,stylegan2} are supposed to reach an equilibrium between $G$ and $D$, this is barely fulfilled in practice.
Typically, $D$ wins over $G$ most of the time.
In this section, we attempt to investigate the cause of such a phenomenon.
\cref{subsec:disequilibrium} briefly reviews the formulation of GANs and describes the disequilibrium between $G$ and $D$.
\cref{subsec:spatial-awareness} interprets the visual attention learned by $D$, which could be one of the key sources for the disequilibrium.

\subsection{Learning Objective of GAN}\label{subsec:disequilibrium}

GAN training~\cite{goodfellow2014generative} is formulated as a two-player game, where a generator is trained to recover the real distribution $p_r$ over data $\{\mathbf{x}\}$, while a discriminator is optimized to differentiate between $p_r$ and the generated distribution $p_g$.
The overall objective function is 
\begin{equation}
\label{eq:minimax}
\begin{split}
\min_G \max_D V(D, G) =&\ \mathbb{E}_{\mathbf{x} \sim p_r}[\log D(\mathbf{x})] \\ 
                      +&\ \mathbb{E}_{\mathbf{z} \sim p_z}[\log (1 - D(G(\mathbf{z})))],
\end{split}
\end{equation}
where $\mathbf{z}$ is a randomly sampled vector, subject to a prior distribution, $p_z$.
In particular, Goodfellow~\textit{et al.}~\cite{goodfellow2014generative} point out that the minimax game defined by GAN will reach an equilibrium point, 
where $G$ recovers the training data distribution while $D$ fails to distinguish the real and fake distributions.
In that case, $D$ is supposed to assign same realness scores to the real and synthesized samples.

\vspace{2pt}
\noindent \textbf{Observation of Disequilibrium.} 
We observe that such an equilibrium is seldom achieved in most GAN variants~\cite{bigGAN, stylegan, stylegan2}.
It appears that $D$ almost always dominates the $G$-$D$ competition, assigning a higher score to real data.
An example of training StyleGAN2~\cite{stylegan2} on LSUN Cat~\cite{yu2015lsun} is shown in \cref{Fig:Logits_Cat}.
We can spot an obvious gap even between $D$'s minimum score for real images and the maximum score for synthesized images.
Such a phenomenon implies that the generated distribution $p_g$ is far from the real distribution $p_r$, affecting the synthesis quality.

\subsection{Visual Attention of Discriminator}\label{subsec:spatial-awareness}

%
After observing the disequilibrium in the interplay between $D$ and $G$, we would like to investigate the behavior of $D$ to see how it manages to outperform $G$.
Prior work on network interpretability, like CAM~\cite{cam}, has found that a classifier tends to focus on some discriminative regions to categorize a given image to the proper class.
However, the discriminator in GANs is trained with the relatively weak supervision, \textit{i.e.}, only having real or fake labels.
Whether it can learn the attentive property from such a bi-classification task remains unknown.
To look under the hood, we apply GradCAM~\cite{gradcam} as an interpretability tool on the well-trained discriminator of StyleGAN2~\cite{stylegan2}. 

Specifically, for a certain layer and a certain class, GradCAM calculates the importance weight of each neuron by average-pooling the gradients back-propagated from the final classification score, over the width and height. 
It then computes the attention map as a weighted combination of the importance weight and the forward activation maps, followed by a ReLU~\cite{glorot2011deep} activation.
The attention map has the same spatial shape as the corresponding feature map.
In this work, we report the GradCAM attention maps all using gradients computed via maximizing the output of $D$.
It reflects the spatial preference of $D$ in making a `real' decision.
In practice we find the attention maps are almost the same if instead minimizing the output of $D$, which indicates the areas that largely contribute to the decision are the same for a discriminator, no matter positively or negatively.
The region with higher response within the attention map contributes more to the decision.
%

\cref{Fig:Hie} visualizes some GradCAM results under multiple feature resolutions.
They are obtained from the discriminators of two StyleGAN2 models, trained on LSUN Cat and FFHQ respectively.
We have following observations:
(1) $D$ learns its own visual attention on both real and generated images.
It suggests that $D$ makes the real/fake decision by paying more attention to some particular regions.
(2) The visual attention emerging from $D$ shows a hierarchical property.
In the shallow layers (like $64\times64$ and $32\times32$ resolutions), $D$ is attentive to local structures such as edge lines in the image.
As the layer goes deeper, $D$ progressively concentrates on the overall location of discriminative contents, \textit{e.g.}, the face of a cat.
(3) The hierarchical attention maps have fewer `local peaks' at more abstract feature layers with a lower resolution.
For example, there is only one peak in the $4\times4$ attention maps.
%

\section{Improving GAN Equilibrium}\label{sec:method}

As shown in \cref{sec:analysis}, the discriminator of GANs has its own visual attention when determining real or fake image.
However, when learning to transform a latent vector into a realistic image, the generator receives no explicit clue about which regions to focus on.
Specifically, for a particular synthesis, $G$ has to decode all the needed information from the input latent code.
Furthermore, $G$ has no idea about the spatial preference of $D$ on making the real/fake decisions.
Such an information asymmetry puts $G$ at a disadvantage when competing with $D$.
In this section, we propose to raise the spatial awareness of $G$ to lessen the information gap between $G$ and $D$.
The overall framework is illustrated in \cref{Fig:Landscape}, which mainly consists of two steps, (1) explicitly encoding spatial awareness into $G$ with a hierarchical heatmap sampling strategy and (2) aligning the spatial awareness of $G$ with the visual attention from $D$ via a feedback regularizer.
\cref{subsec:spatial-awareness-to-g} and \cref{subsec:feedback-mechanism} introduce these two techniques respectively.

\subsection{Encoding Spatial Awareness in Generator}\label{subsec:spatial-awareness-to-g}

\vspace{2pt}
\noindent \textbf{Hierarchical Heatmap Sampling.}
To improve the awareness of $G$ on spatial regions, we propose a hierarchical heatmap sampling algorithm.
This heatmap is responsible for teaching $G$ which regions to pay more attention to.
Inspired by the visual attention induced from $D$ as in \cref{subsec:spatial-awareness}, we abstract our heatmap as a combination of several sub-regions and a background.
Taking the heatmap at the $4\times4$ resolution as an example (leftest in \cref{Fig:Landscape}), it tells $G$ there is one region to focus on, whose center locates at the black dot.
We formulate each sub-region as a 2D map, $H_i$, which is sampled subject to a Gaussian distribution
\begin{equation}
    H_i \sim \mathcal{N} (\mathbf{c_i}, \mathbf{cov}), \label{eq:heatmap-sampling}
\end{equation}
where $\mathbf{c_i}$ and $\mathbf{cov}$ denote the mean and the covariance.
According to the definition of 2D Gaussian distribution, $\mathbf{c_i}$ just represents the coordinates of the region center.
The final heatmap can be written as the sum of all sub-maps, $H=\sum_{i=1}^{n}H_i$, where $n$ denotes the total number of local regions for $G$ to focus on.

As pointed out in the prior works~\cite{stylegan, higan}, the generator in GANs learns image synthesis in a coarse-to-fine manner, where the early layers provide a rough template and the latter layers refine the details.
To match such a mechanism, we design a hierarchical heatmap sampling algorithm.
Concretely, we first sample a spatial heatmap with \cref{eq:heatmap-sampling} for the most abstract level (\textit{i.e.}, with the lowest resolution), and derive the heatmaps for other resolutions based on the initial one.
The number of centers, $n$, and the covariance, $\mathbf{cov}$, adapt accordingly to the feature resolution.

\begin{figure}[tbp]
\begin{center}
\includegraphics[width=0.9\linewidth]{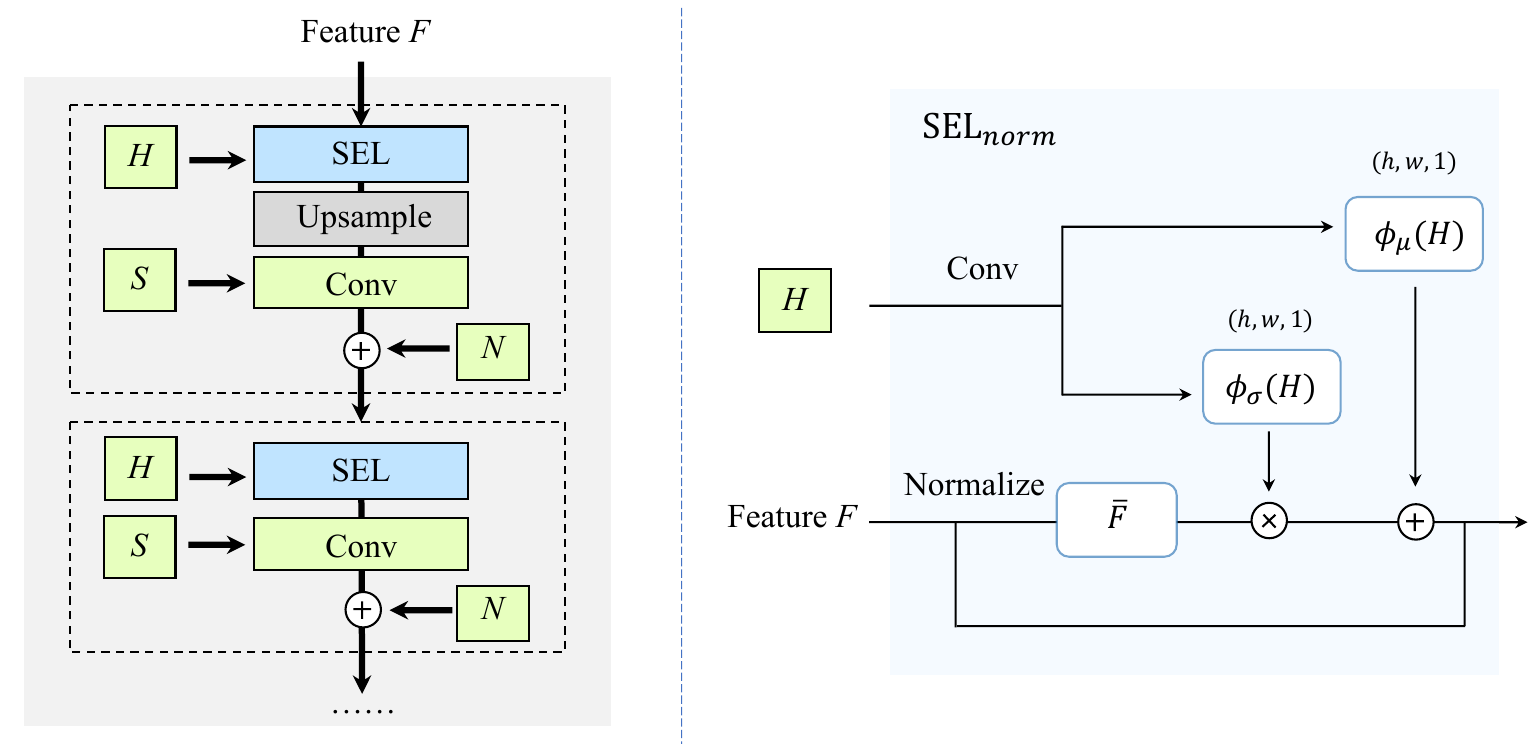}
\vspace{-5pt}
  \caption{\textbf{Spatial Encoding Layer.} The left shows how the layer works over StyleGAN2 at each resolution, and the right describes the internal of the SEL$_{\textit{norm}}$. The symbol `S' represents the style in StyleGAN2, `N' is the noise, and `H' indicates the spatial heatmap. Learning from ~\cite{spade}, we incorporate the spatial heatmaps into $G$ via normalization and denormalization. 
  }
\label{Fig:SEL}
\vspace{-15pt}
\end{center}
\end{figure}

\vspace{2pt}
\noindent \textbf{Heatmap Encoding.}
We incorporate the spatial heatmaps into $G$ to raise its spatial awareness.
It generally can be conducted in two ways, via feature concatenation or feature normalization~\cite{adain,spade}.
We use a spatial encoding layer (SEL), respectively trying these two variants, denoted as SEL$_{\textit{concat}}$ and SEL$_{\textit{norm}}$.
Specifically, inspired by SPADE~\cite{spade}, the variant SEL$_{\textit{norm}}$ integrates the hierarchical heatmaps into the per-layer feature maps of $G$ with normalization and denormalization operations, as
\begin{equation}
    SEL_{\textit{norm}}(F,H) = \phi_{\sigma}(H) \  \frac{F- \mu(F)}{\sigma(F)} + \phi_{\mu}(H),
\end{equation}
where $F$ denotes an intermediate feature map produced by $G$, which is with the same resolution as $H$.
$\mu(\cdot)$ and $\sigma(\cdot)$ respectively stands for the functions of computing channel-wise mean and standard deviation.
%
%
%
$\phi_{\mu}(\cdot)$ and $\phi_{\sigma}(\cdot)$ are two learnable functions, whose outputs are point-wise and with a shape of $(h, w, 1)$.
%
%
Besides, as shown in \cref{Fig:SEL}, we use a residual connection to stabilize the intermediate features.
If not particularly specified, this paper adopts the variant SEL$_{\textit{norm}}$ since it shows a slightly better performance.  
%

It is worth noting, although we learn the SEL$_{\textit{norm}}$ architecture from SPADE~\cite{spade}, these two methods are clearly different since SPADE targets at synthesizing images based on a given semantic segmentation mask, whose training requires paired ground-truth data, while our model is trained with completely unlabeled data. Meanwhile, SEL$_{\textit{norm}}$ is just a replaceable component of our approach.
%
%


\subsection{Aligning Spatial Awareness with Discriminator}\label{subsec:feedback-mechanism}

Encoding heatmaps into $G$ can explicitly raise its spatial awareness, but it is not enough to make $G$ competitive with $D$.
The reason is that, $D$ learns its own visual attention based on the semantically meaningful image contents, but the heatmaps fed into $G$ are completely arbitrary.
Without further guidance, how $G$ is supposed to utilize the heatmaps is ambiguous.
For example, $G$ has no idea about ``whether to pay \textit{more} or \textit{less} attention on the highlighted regions in the heatmap''.
To make the best usage of the introduced spatial awareness, we propose to involve $D$ as a regularizer to supervise $G$, to properly leverage the spatial knowledge.

Specifically, at each optimization step of $G$, we use $D$ to generate the visual attention map via GradCAM as a self-supervision signal.
Besides competing with $D$, $G$ is further trained to minimize the distance between the attention map induced from $D$ and the input heatmap $H$.
The loss function can be written as
\begin{align}  
\mathcal{L}_{\textit{align}} = || \  \text{GradCAM}_{D}[G(H, \mathbf{z})] - \ H \ ||_1.
\end{align} 
We truncate the $\mathcal{L}_{\textit{align}} $ values if smaller than a constant $\tau$, since the sampled heatmaps are not expected to perfectly match the real attention maps shaped by semantics.
The threshold $\tau$ is set as $0.25$ for all the experiments.
Note that $D$ is not updated in the process above and only used as a supervision signal to train $G$.
Such a regularization loss aligns the spatial awareness of $G$ with the spatial preference of $D$, narrowing the information gap between them.

\section{Experiments}
\label{sec:exp}

\begin{figure*}[t]
    \centering
    \includegraphics[width=0.95\linewidth]{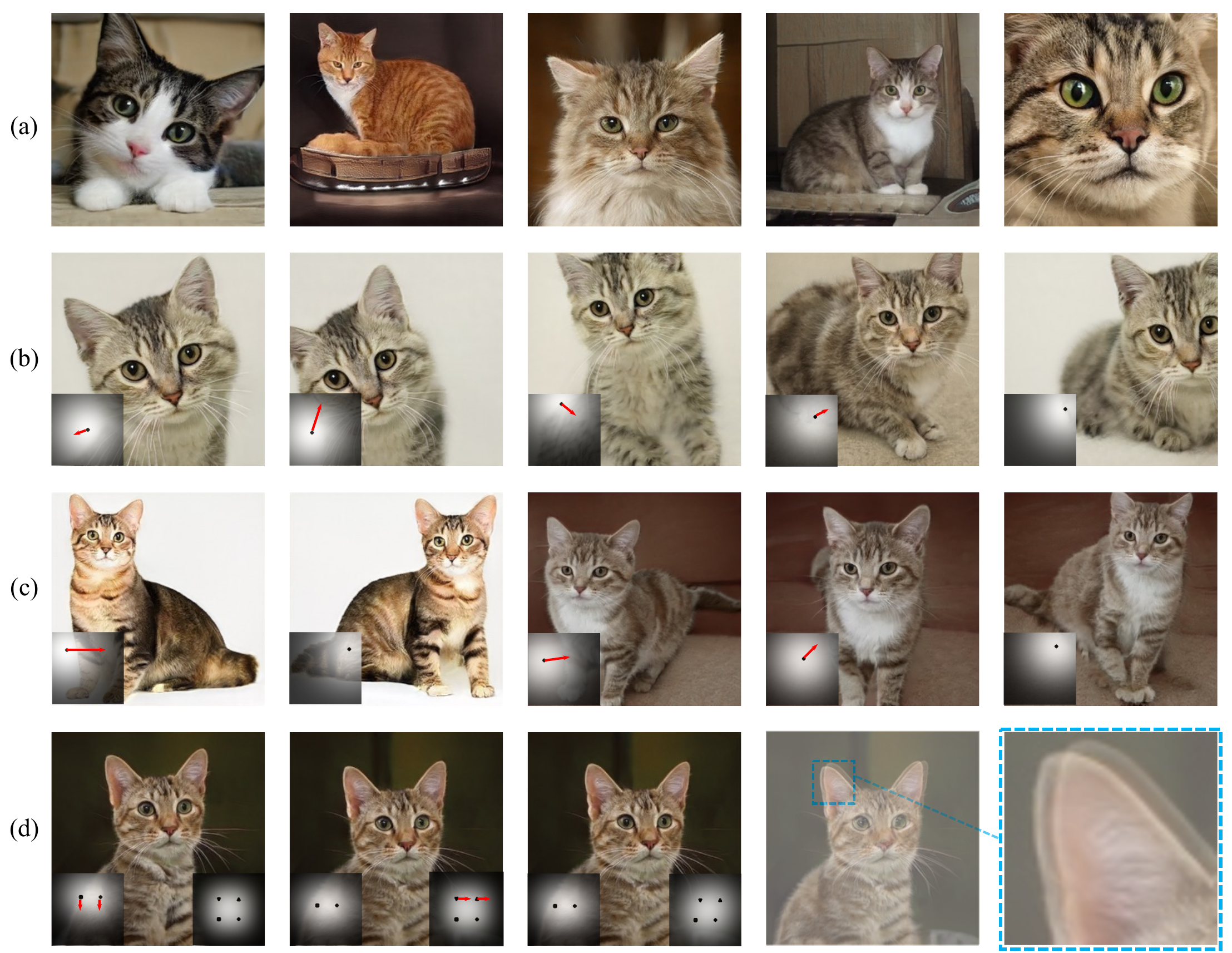}
    \vspace{-8pt}
    \caption{\textbf{Qualitative results on LSUN Cat dataset and the demonstration of spatial awareness via varying the spatial heatmaps of the generator.} Row (a) shows several generated samples of a model trained through EqGAN-SA.
    Rows (b) and (c) illustrate the spatial awareness of $G$: we keep the latent codes unchanged and move the spatial heatmap at the $4\times4$ level. The arrows indicate the movement direction, where the cat moves along with the varied heatmap. To further show the hierarchical structure, we move the heatmap at the finer level in the Row (d). Different from the body movement, the change in $8\times8$ heatmap (two centers) mainly moves the cat eyes, and the change in $16\times16$ heatmap (four centers) leads to subtle movement of the cat ears. It is worth noting that, as the content is being manipulated, our $G$ knows to adjust the nearby regions to make everything coherent. 
    }
    \label{Fig:Cat_Spatial_Awareness}
    \vspace{-5pt}
\end{figure*}

\begin{figure*}[htbp]
    \centering
    \includegraphics[width=0.95\linewidth]{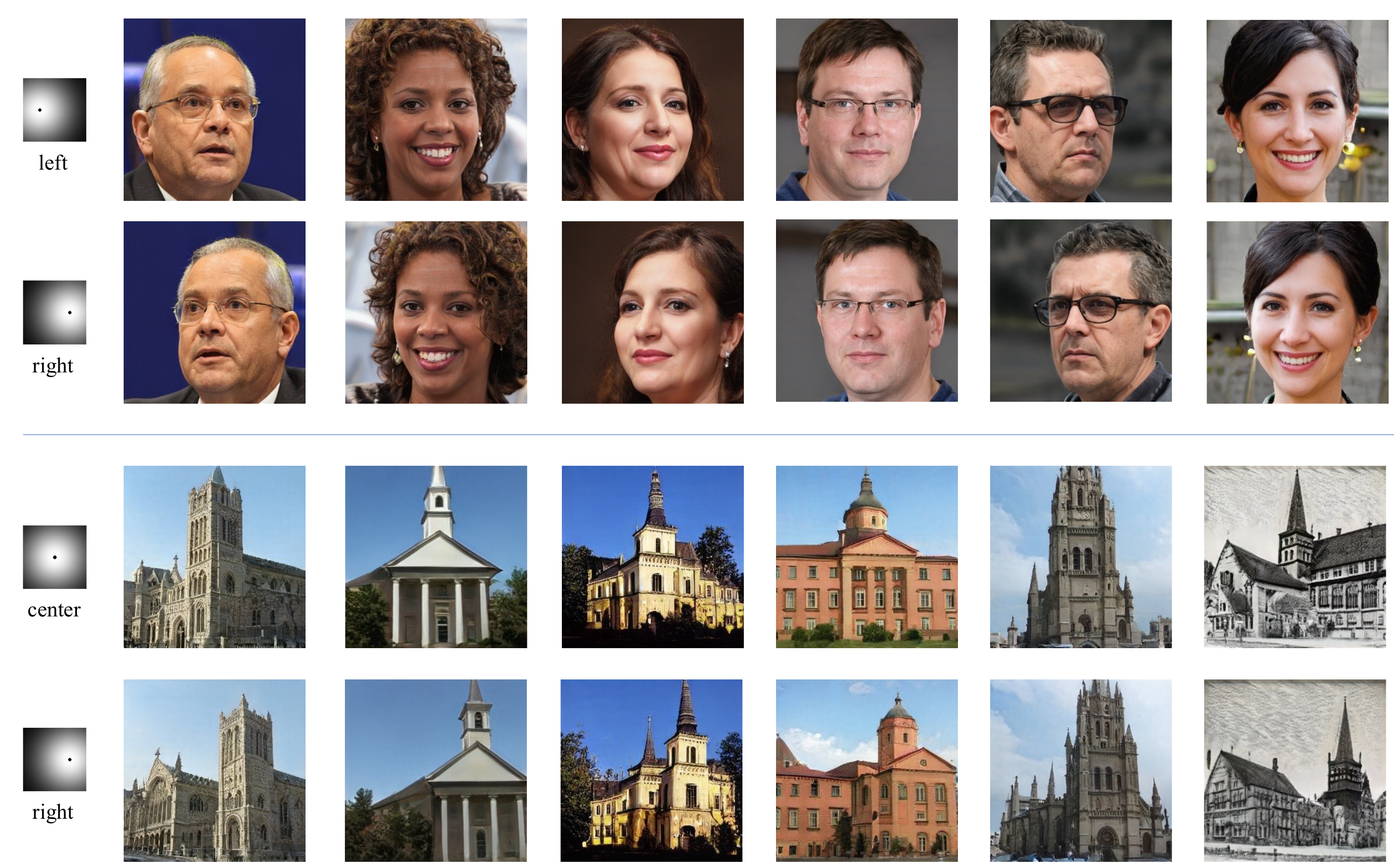}
    \vspace{-8pt}
    \caption{\textbf{Qualitative results on the FFHQ dataset (top) and the LSUN Church dataset (bottom)}. Each row uses the same spatial heatmap but different latent codes, and each column uses the same latent code. We can see that the spatial heatmap roughly controls the pose of the face and the viewpoint of the church building, which facilitates the interactive spatial editing of the output image. 
    } 
    \label{Fig:FFHQ_Spatial_Awareness}
    \vspace{-5pt}
\end{figure*}

%
We evaluate the proposed EqGAN-SA on multiple benchmarks. Sec.~\ref{subsec:impl} provides the implementation details. The main comparison and experimental results are presented in Sec.~\ref{subsec:main-results}. Our EqGAN-SA could improve the spatial attentive property in $G$ and mitigates the disequilibrium to some extent. Sec.~\ref{subsec:ablation} includes the comprehensive ablation studies on the role of each proposed component.

\subsection{Implementation Details}\label{subsec:impl}

\noindent \textbf{Datasets.} We conduct the experiments on the LSUN Cat~\cite{yu2015lsun}, FFHQ~\cite{stylegan}, and LSUN Church~\cite{yu2015lsun} datasets.
The LSUN Cat dataset contains $1600K$ real-world images regarding different cats.
Following the setting of ~\cite{stylegan2-ada}, we take $200K$ image samples from the LSUN Cat dataset for training.
The FFHQ dataset consists of $70K$ high-resolution ($1024\times1024$) images of human faces, under Creative Commons BY-NC-SA 4.0 license~\cite{FFHQ_github}. Usually, the images are horizontally flipped to double the size of training samples. 
The LSUN Church dataset includes $126K$ images with visually complex church scenes. It is worthy noting that all images are resized to $256 \times 256$ resolution.

\noindent \textbf{Spatial Heatmap Sampling and Encoding.}
In practice, we find the GradCAM maps on the fine resolutions are too sensitive to semantic cues.
Therefore, we only conduct encoding on the level $0,1,2$ of $G$, \emph{i.e.}, resolution $4\times4$, $8\times8$, and $16\times16$.
We heuristically generate $1,2,4$ centers (in other words, sub-heatmaps) on these three levels.
We sample the level $0$ heatmap center $\mathbf{c^{0}_0}$ by a Gaussian distribution with a mean of $(\frac{h}{2},\frac{w}{2})$, and a standard deviation of $(\frac{h}{3},\frac{w}{3})$.
To keep the heatmaps consistent at various levels, we sample the level $1,2$ centers over the level $0$ center.
It indicates the mean of Gaussian distribution $\mathbf{c^{1}_n}$ and $\mathbf{c^{2}_n}$ is the sampled $\mathbf{c^{0}_0}$.
Their standard deviations are $(\frac{h}{6},\frac{w}{6})$.
If we shift the level $0$ center, the heatmaps of other levels will move correspondingly.
Following the coarse-to-fine manner, we decrease each center's influence area level by level.
Besides, we drop the sampling if the level $0$ center is outside the image.
In our observation, the results of the proposed method are robust to these hyperparameters for heatmap sampling.
Therefore, we use the same hyperparameters for heatmap sampling on all the datasets.
More implementation details are provided in \textit{Supplementary Material}.

\noindent \textbf{Training.} We implement our EqGAN-SA on the official implementation of \href{https://github.com/NVlabs/stylegan2-ada-pytorch}{StyleGAN2}, such that the state-of-the-art image generation method StyleGAN2~\cite{stylegan2} serves as our baseline.
We follow the default training configuration of \cite{stylegan2-ada} for the convenience of reproducibility, and keep the hyperparameters unchanged to validate the effectiveness of our proposed framework.
For example, we train all the models with a batch size of $64$ on $8$ GPUs and continue the training until $25M$ images have been shown to the discriminator. 
Our method increases the training time by around $30\%$ compared with the baseline.

\noindent \textbf{Evaluation.}
We use Fréchet Inception Distance (FID)~\cite{heusel2017gans} between $50K$ generated samples and all the available real samples as the image generation quality indicator.
We utilize a specific approximation of Wasserstein distance to quantify the degree of disequilibrium, \ie, the distance between (a) the minimum discriminator scores for real samples and (b) the maximum scores for generated samples.
We term it as Disequilibrium Indicator (DI), where $\text{DI}=\min(s_r)-\max (s_g)$.
It indicates if $D$ can distinguish between the hardest real image and most realistic generated image.
To keep the result stable, we compute DI over $128$ randomly sampled images ($64$ real and $64$ fake) for $200$ times and take the mean value.
%
We also discuss the validity of DI in \textit{Supplementary Material}.
\subsection{Main Results}\label{subsec:main-results}

\noindent \textbf{Spatial Awareness is Raised in Generator}. 
As discussed in \cref{subsec:spatial-awareness-to-g}, we propose to encode spatial awareness into $G$. 
Here we provide the qualitative results in ~\cref{Fig:Cat_Spatial_Awareness} and ~\cref{Fig:FFHQ_Spatial_Awareness} to verify that $G$ indeed learns to focus on the regions specified by input heatmaps.
Specifically, we keep the latent vector unchanged and move the spatial heatmaps. 
As we move the level $0$ heatmap of the sample (b) and (c) in ~\cref{Fig:Cat_Spatial_Awareness}, the cat bodies move under the guidance of heatmap movement (indicated by red arrows).
We can observe the same phenomenon by watching each column of ~\cref{Fig:FFHQ_Spatial_Awareness}.
Furthermore, as illustrated by each row of ~\cref{Fig:FFHQ_Spatial_Awareness}, the human images generated with the same heatmap will put faces on the same location.
In addition, as desired by our hierarchical design, moving level $1$ and $2$ heatmaps would affect local structures.
For example, in the sample (d) of ~\cref{Fig:Cat_Spatial_Awareness}, the change in level $1$ heatmap leads to a movement in cat eyes.
As we slightly push the top two centers of level $2$ heatmap to the right, the cat ears subtly turn right while other parts, even the cat whiskers, remain unchanged.
These verify the effect of our hierarchical spatial encoding.
We also notice $G$ could adaptively modify the nearby texture and structure to give a reasonable image.
Additionally, we visualize the generator intermediate features to investigate whether it has spatial awareness and the effect of our method, as shown in \textit{Supplementary Material}.
Overall, the moved contents depict the spatial awareness of $G$, which shows a hierarchical style and matches our design target.

\noindent \textbf{Equilibrium is Improved.}
The quantitative results on the three datasets are provided in ~\cref{tab:main_results}.
On all the datasets, the metric $\text{DI}$ shows a drop after encoding spatial awareness into $G$, and a further decrease with the help of  $\mathcal{L}_{\textit{align}}$.
For example, $\text{DI}$ reduces from $3.64$ to $3.12$ and finally $2.39$ on the LSUN Cat dataset.
This observation verifies the hypothesis that the aforementioned information asymmetry is a source of GAN disequilibrium, and our proposed approach can mitigate the imbalance. 

With the improved equilibrium, the image synthesis quality also becomes better. We observe that there are consistent improvements over the FID on three datasets, outperforming the baseline StyleGAN2. 
We also validate our idea on the basis of SN-DCGAN (DCGAN~\cite{dcgan} with spectral normalization~\cite{miyato2018spectral}) on the CIFAR-10~\cite{cifar10} dataset, as shown in ~\cref{tab:sndcgan}. 

\begin{table}[t]
\small
\begin{center}
\caption{\textbf{Quantitative results on LSUN Cat, FFHQ, and LSUN Church datasets, all trained with $\mathbf{25}\mathbf{M}$ images shown to discriminator.} The baseline uses the architecture of StyleGAN2~\cite{stylegan}. We use FID as the metric for image generation quality. We also formulate a metric Disequilibrium Indicator (DI), a specific form of Wasserstein distance~\cite{wgan}, to quantify the disequilibrium. DI is calculated as $\min(s_r)-\max(s_g)$, where $s$ indicates the discriminator outputs before the activation. We discuss the validity of DI and include the results of other metrics in \textit{Supplementary Material}.
%
$\downarrow$ denotes smaller is better. }
\vspace{-5pt}
\begin{tabular}{c|cc |cc|cc}
\hline
 \multirow{3}{*}{Method}  & \multicolumn{2}{c|}{Cat~\cite{yu2015lsun}} & \multicolumn{2}{c|}{FFHQ~\cite{stylegan}} & \multicolumn{2}{c}{Church~\cite{stylegan}} \\
 & \multicolumn{2}{c|}{$256\times256$} & \multicolumn{2}{c|}{$256\times256$} & \multicolumn{2}{c}{$256\times256$} \\
\cline{2-7}
& FID $\downarrow$ & DI $\downarrow$ & FID $\downarrow$ & DI $\downarrow$ & FID $\downarrow$ & DI $\downarrow$  \\
\hline
Baseline & 8.36 &  3.64  & 3.66 & 1.62 & 3.73 &3.01  \\
+ SEL & 7.82 & 3.12  & 3.39 & 1.38 & 3.55& 2.59 \\
+ $\mathcal{L}_{\textit{align}}$ & \textbf{6.81} & \textbf{2.39} & \textbf{2.96} & \textbf{0.73} & \textbf{3.11} & \textbf{2.07} \\
\hline
\end{tabular}
\label{tab:main_results}
\end{center}
\vspace{-10pt}
\end{table}

\subsection{Ablation Study}\label{subsec:ablation}
\noindent \textbf{How Important is the Type of Spatial Heatmap Sampling?} 
Different sampling strategies are applied here to validate our choice, as provided in ~\cref{tab:ab_heatmap}.
Specifically, 2D Gaussian noise is first considered as a straightforward baseline experiment since it provides non-structured spatial information. Accordingly, 2D Gaussian noise introduces no performance gains. It indicates, merely feeding a 2D heatmap but without any region to be emphasized is insufficient to raise spatial awareness and mitigate the disequilibrium.

Besides, we also use multiple-resolution spatial heatmaps but discard the hierarchical constraint, referred as Non-Hie in~\cref{tab:ab_heatmap}. Namely, spatial heatmaps at different resolutions are independently sampled. Obviously, the baseline is improved by this non-hierarchical spatial heatmap, demonstrating the effectiveness of the spatial awareness of $G$. Moreover, when the hierarchical sampling is adopted, we observe further improvements over the synthesis quality and equilibrium.

\noindent \textbf{How Important is the Way of Spatial Encoding?} 
In order to raise the spatial awareness of $G$, there exist several alternatives to implement. Therefore, we conduct an ablation study on LSUN Cat and FFHQ datasets to test various methods. 
For example, the first way of feeding the spatial heatmap is to flatten the 2D heatmap as a vector, and then concatenate it with the original latent code. This setting aims at validating whether maintaining 2D structure of spatial heatmap is necessary. Besides, we also use two different SEL modules (\emph{i.e.}, SEL$_{\textit{concat}}$ and SEL$_{\textit{norm}}$) mentioned in Sec.~\ref{subsec:spatial-awareness-to-g}. Their details are available in \textit{Supplementary Material}. For a fair comparison, all the ablation studies use $\mathcal{L}_{\textit{align}}$.

\cref{tab:ab_spatial} presents the results. 
Apparently, simply feeding the spatial heatmap but without the explicit $2D$ structure leads to no gains compared to the baseline. It might imply that it is challenging to use a vector (like the original latent code) to raise the spatial awareness of the generator. Instead, the proposed SEL module could introduce the substantial improvements, demonstrating the effectiveness of the encoding implementation.

\begin{table}[t]
    \centering
    \caption{
    \textbf{Quantitative results on the CIFAR-10 dataset over the baseline SN-DCGAN,} with conditional or  unconditional image synthesis.
    }
    \vspace{-5pt}
    \small
    \setlength{\tabcolsep}{10pt}
    \begin{tabular}{c|c|c|c|c}
        \hline
        \multirow{2}{*}{Method} & \multicolumn{2}{c|}{Unconditional} & \multicolumn{2}{c}{Conditional} \\
        \cline{2-5}
              & FID $\downarrow$ &  DI $\downarrow$                   & FID $\downarrow$  &  DI $\downarrow$ \\
        \hline
        SN-DCGAN &    23.72 & 1.85                   &    19.89      &     1.61  \\
       + Ours  &      \textbf{16.93}  &  \textbf{0.96 }                  &     \textbf{13.56}     &   \textbf{0.78}    \\
        \hline
    \end{tabular}
    \label{tab:sndcgan}
    \vspace{-5pt}
\end{table}

\begin{table}[t]
\small
\begin{center}
\caption{\textbf{Spatial Heatmap Sampling.} With other parts unchanged, we separately throw random Gaussian noise, spatial heatmap without hierarchical sampling, and our spatial heatmap as the input to the spatial encoding layer.}
\vspace{-5pt}
\setlength{\tabcolsep}{9pt}
\begin{tabular}{c|c|c|c|c}
\hline
& Baseline & Gau. Noise & Non-Hie & Hie  \\
\hline
FID $\downarrow$ & 8.36&8.31& 7.29 & \textbf{6.81}\\
DI $\downarrow$ &3.64&3.67& 2.70 & \textbf{2.39}\\
\hline
\end{tabular}
\label{tab:ab_heatmap}
\end{center}
\vspace{-10pt}
\end{table}

\begin{table}[t]
\small
\begin{center}
\caption{\small \textbf{Ablation study on spatial encoding.} We flatten the spatial heatmap and incorporate the vectorized one into latent code, denoted as `Flatten'. It destroys the $2D$ space structure, and hence cannot improve over the baseline. Instead, encoding heatmaps in the spatial domain is beneficial. The two variants of SEL show a similar result, where SEL$_{\textit{norm}}$ is slightly better. }
\vspace{-5pt}
\setlength{\tabcolsep}{12pt}
\begin{tabular}{c|c|c|c|c}
\hline
 \multirow{2}{*}{Method}  & \multicolumn{2}{c|}{Cat~\cite{yu2015lsun}} & \multicolumn{2}{c}{FFHQ~\cite{stylegan}} \\
\cline{2-5}
 & FID $\downarrow$ & DI $\downarrow$ & FID $\downarrow$ & DI $\downarrow$ \\
\hline
Baseline & 8.36 & 3.64 &3.66&1.62 \\
Flatten  & 8.63 & 3.71 & 3.78 &1.63  \\
SEL$_{\textit{concat}}$ & 7.02 & 2.47 & 3.11 & 0.90  \\
SEL$_{\textit{norm}}$  & \textbf{6.81} & \textbf{2.39} & \textbf{2.96} & \textbf{0.73} \\
\hline
\end{tabular}
\label{tab:ab_spatial}
\end{center}
\vspace{-10pt}
\end{table}

\begin{figure}[t]
\begin{center}
\includegraphics[width=\linewidth]{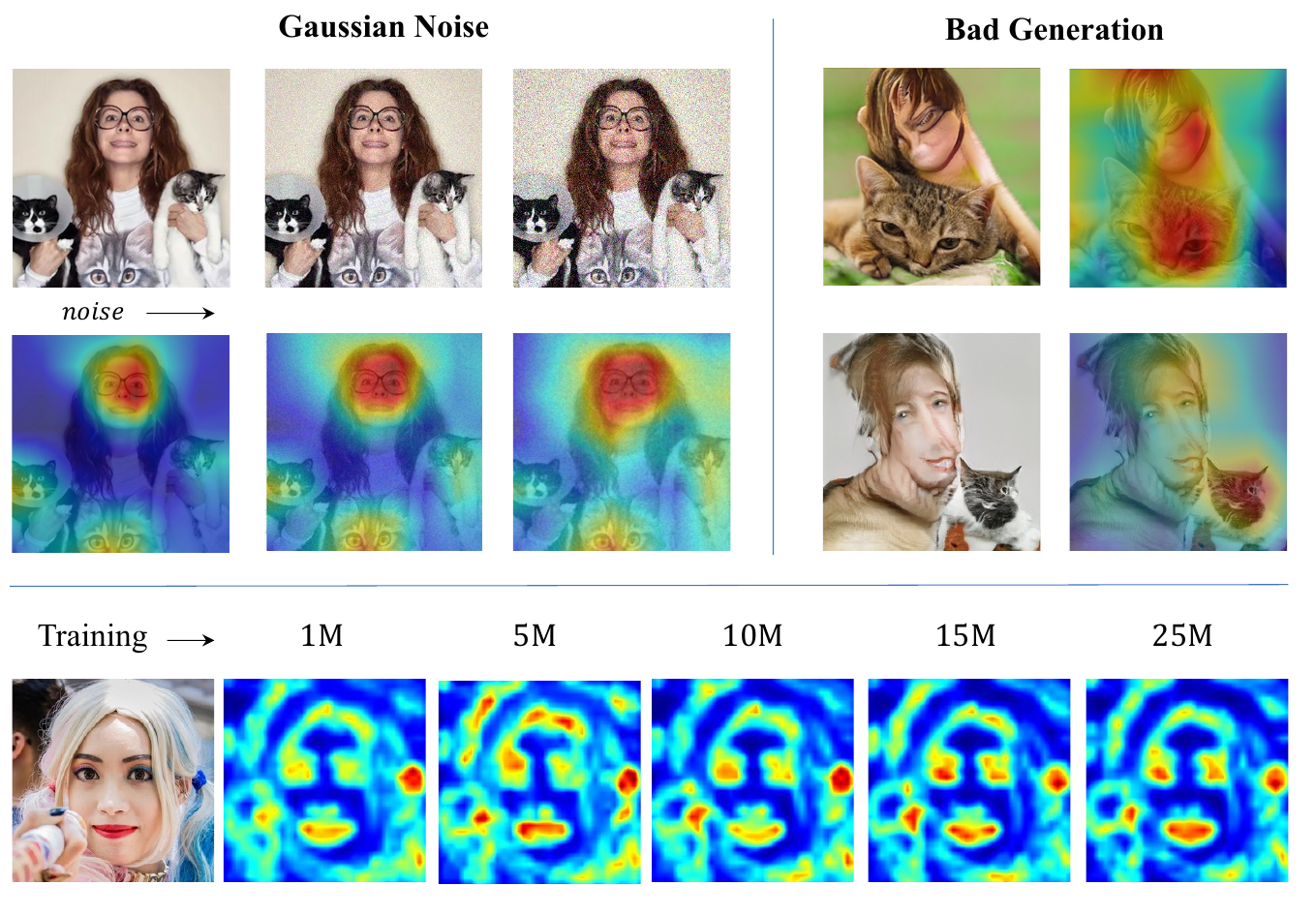}
    \vspace{-25pt}
  \caption{\textbf{Robustness and Consistency.} We test the response of $D$ to noisy images and bad generation samples in the top. The bottom visualizes that the $D$'s attention is consistent over the training. 
  }    
\label{Fig:Robustness}
\end{center}
\vspace{-10pt}
\end{figure}

\noindent \textbf{Whether Visual Attention of $D$ is Robust and Consistent?}
As discussed in ~\cref{subsec:feedback-mechanism}, the alignment loss ($\mathcal{L}_{\textit{align}}$) uses the  $D$ attention maps to guide $G$.
It assumes the attention map from $D$ is stable enough to serve as a supervision signal and valid over the whole training.
To validate the design, we first explore the \emph{robustness} of $D$.
As shown in the left top of ~\cref{Fig:Robustness}, we add random Gaussian noise to a real image from the LSUN Cat dataset, destroying its texture.
As the noise amplitude increasing, we can visually see the noise pattern and the local appearance has been over smoothed.
$D$ is still attentive to the original important regions, \emph{e.g.}, the human and cat faces.
We then test its response to terrible samples generated by a poorly-trained $G$, illustrated in the right top of ~\cref{Fig:Robustness}.
The samples contain distorted human, cat and background.
That is, the visual attention of $D$ is sufficiently robust to the noise perturbation and the generated artifacts. 
Furthermore, as indicated in the bottom of ~\cref{Fig:Robustness}, we validate whether the visual attention is \emph{consistent} throughout the entire training process. 
At a very early stage of training, $D$ has already localized the discriminative regions. The focus of such visual attention is consistently maintained till the end of the training. 
The robustness and consistency property of $D$ attention could successfully provide a support for $\mathcal{L}_{\textit{align}}$.

\section{Discussion}

\noindent \textbf{Related Work.} Generative adversarial networks (GANs) \cite{goodfellow2014generative} have shown a great success in many generative tasks, such as synthesising photorealistic images.
It aims to recover the target distribution via a minimax two-player game, whose global optimum exists as a Nash Equilibrium~\cite{goodfellow2014generative,heusel2017gans,fedus2017many,pmlr-v97-hsieh19b}.
Researchers have developed numerous techniques to improve the synthesis quality of GANs, through a Laplacian pyramid framework~\cite{denton2015deep}, an all-convolutional deep neural network~\cite{dcgan}, progressive training~\cite{pggan}, spectral normalization~\cite{miyato2018spectral,zhang2019self}, and large-sacle GAN training~\cite{bigGAN,BigBiGAN}.
Recently, the style-based architecture StyleGAN~\cite{stylegan} and StyleGAN2~\cite{stylegan2} have become the state-of-the-art method for image synthesis, by separating high-level attributes.
Besides, some methods also incorporate additional information into discriminator or generator, such as pixel-wise representation~\cite{unetgan}, 3D pose~\cite{giraffe}, or neighboring instances~\cite{instanceGAN}.

In the early development stage, some methods study the equilibrium between $G$ and $D$ to stabilize training and enhance the synthesis quality \cite{wgan,fedus2017many,berthelot2017began}.
However, the equilibrium problem seems to be neglected in recent years, possibly covered up by the great success in other aspects like the architecture design. 
Instead, we verify that improving GAN equilibrium could lead to a substantial performance gain, even on the state-of-the-art method StyleGAN2. 
%


\noindent \textbf{Limitation.}
Though simple and effective, our EqGAN-SA is heuristic and built upon existing techniques. 
In addition, we notice the spatial encoding operation would sometimes lead to a synthesis blurring at the location of heatmaps boundaries.
We consider EqGAN-SA as an empirical study to show that the asymmetry between the spatial awareness of $G$ and $D$ is a source of the GAN disequilibrium.
We hope this work can inspire more works of revisiting the GAN equilibrium and develop more novel methods to improve the image synthesis quality through maneuvering the GAN equilibrium. We will also conduct more theoretical investigation on this issue in the future work.

\noindent \textbf{Ethical Consideration.}
This paper focuses on studying the disequilibrium of GANs to improve the image synthesis quality.
Although only using the public datasets for research and follow their licences, the abuse of our method may bring negative impacts through deep fake generation.
Such risks would increase as the synthesis results of GANs are becoming more and more realistic.
From the perspective of academia, these risks may be mitigated by promoting the research on deep fake detection.
It also requires the management on the models trained with sensitive data.

\section{Conclusion}
\label{sec:con}
In this paper we explore the problem of GAN equilibrium, and identify one of its possible attributing sources is the information asymmetry between $G$ and $D$.
Specifically, we notice that $D$ spontaneously learns its visual attention while $G$ is not aware of which spatial regions to focus on for a particular synthesis.
Therefore, we propose a new training technique EqGAN-SA to reduce such information asymmetry, by enabling spatial awareness of $G$ and aligning it with the attention of $D$.
Qualitative results show that our method successfully makes $G$ to concentrate on specific regions.
Experiments on various datasets validate that our method mitigates the disequilibrium in GAN training and substantially improves the overall image synthesis quality. The resulting model with spatial awareness also enables the interactive manipulation of the output image. 
%

{\small
\bibliographystyle{ieee_fullname}
\bibliography{ref}
}



\end{document}